\documentclass[sigconf]{acmart}
\usepackage[linguistics]{forest}
\usepackage{amsmath}
\usepackage{graphicx,kantlipsum,setspace}
\usepackage{caption}
\usepackage{placeins} % put this in your pre-amble
\usepackage{flafter}  % put this in your pre-amble
\usepackage{enumitem}

\usepackage{cleveref}[2012/02/15]% v0.18.4; 
\crefformat{footnote}{#2\footnotemark[#1]#3}

\acmYear{2020}
\acmDOI{10.1145/nnnnnnn.nnnnnnn}

\acmConference[KDD '20]{KDD '20: Workshop on Applied Data Science for Healthcare}{August 23--27, 2020}{San Diego, CA}

\setcopyright{none} 

\begin{document}

%%
%% The "title" command has an optional parameter,
%% allowing the author to define a "short title" to be used in page headers.
\title{Information Extraction of Clinical Trial Eligibility Criteria}
\author{Yitong Tseo}
\affiliation{%
  \institution{Facebook Inc.}
  \city{Menlo Park}
  \country{USA}}
\email{yitong@fb.com}

\author{M. I. Salkola}
\affiliation{%
  \institution{Facebook Inc.}
  \city{Menlo Park}
  \country{USA}}
\email{salkola@fb.com}

\author{Ahmed Mohamed}
\affiliation{%
  \institution{Facebook Inc.}
  \city{Menlo Park}
  \country{USA}}
\email{ahmedkm@fb.com}

\author{Anuj Kumar}
\affiliation{%
  \institution{Facebook Inc.}
  \city{Menlo Park}
  \country{USA}}
\email{anujk@fb.com}

\author{Freddy Abnousi, MD}
\affiliation{%
  \institution{Facebook Inc.}
  \city{Menlo Park}
  \country{USA}}
\email{abnousi@fb.com}

\begin{abstract}
Clinical trials predicate subject eligibility on a diversity of criteria ranging from patient demographics to food allergies. Trials post their requirements as semantically complex, unstructured free-text. Formalizing trial criteria to a computer-interpretable syntax would facilitate eligibility determination. In this paper, we investigate an information extraction (IE) approach for grounding criteria from trials in ClinicalTrials.gov to a shared knowledge base. We frame the problem as a novel knowledge base population task, and implement a solution combining machine learning and context free grammar (CFG). To our knowledge, this work is the first criteria extraction system to apply attention-based conditional random field architecture for named entity recognition (NER), and word2vec embedding clustering for named entity linking (NEL). We release the resources and core components of our system on GitHub.\footnote{\label{note1}\href{https://github.com/facebookresearch/Clinical-Trial-Parser}{https://github.com/facebookresearch/Clinical-Trial-Parser}} Finally, we report our per module and end to end performances; we conclude that our system is competitive with Criteria2Query, which we view as the current state-of-the-art in criteria extraction \cite{Criteria2Query}.
\end{abstract}

\begin{CCSXML}
<ccs2012>
<concept>
<concept_id>10002951.10003317.10003347.10003352</concept_id>
<concept_desc>Information systems~Information extraction</concept_desc>
<concept_significance>500</concept_significance>
</concept>
<concept>
<concept_id>10010405.10010444.10010447</concept_id>
<concept_desc>Applied computing~Health care information systems</concept_desc>
<concept_significance>300</concept_significance>
</concept>
</ccs2012>
\end{CCSXML}

\ccsdesc[500]{Information systems~Information extraction}
\ccsdesc[300]{Applied computing~Health care information systems}

\maketitle

\section{Introduction}
Clinical trials are vital for understanding diseases and testing new treatments. However trials in the United States today face significant challenges recruiting enough participants \cite{Carlisle05}\cite{Gubar19} and establishing representative diversity in their study populations \cite{Knepper18}, which regularly leads to difficulty completing trials and generalizing outcomes across populations. 
% \looseness=-1

\href{https://clinicaltrials.gov/}{ClinicalTrials.gov} is a centralized public database of 330,000+ clinical studies maintained by the National Library of Medicine (NLM) \cite{ClinicalTrialsdotgov}. In addition to hosting every American and many international trials, ClinicalTrials.gov provides filters for a handful of important eligibility criteria such as patient age, gender, trial location, and study condition. Researchers have the option to specify additional, more specific, eligibility criteria such as treatment history and pre-existing conditions. These criteria are written in free-text descriptions, the majority of which include semantically complex language and can require expert domain knowledge to understand \cite{Ross10}. With 32,000+ new trials added annually to ClinicalTrials.gov, automated criteria extraction is a necessary requisite for sophisticated trial discovery and cohort identification platforms.
% \looseness=-1

Previous work on automated criteria extraction take many approaches \cite{Alves2019InformationEA}\cite{Weng10}. Systems such as EliXR \cite{EliXR}, EliXR-TIME \cite{EliXR-TIME}, and ERGO \cite{ERGO} build on pattern matching and rules. Other researchers such as Butler et al. \cite{Butler18} and Luo et al. \cite{Luo13} create text mining algorithms to identify common criteria across trials. There has also been significant research focusing on information extraction including Bruijn et al.'s work \cite{Bruijn08}, EliIE \cite{EliIE}, and Criteria2Query \cite{Criteria2Query}. 
% \looseness=-1

In this work, we develop an information extraction approach for eligibility criteria extraction which combines machine learning and context free grammar. Our work makes the following contributions: 
% \looseness=-1

\begin{itemize} [itemsep=2pt,parsep=1pt]
\item We formulate eligibility criteria extraction as a novel knowledge base population task. Working from this theoretical framework, we achieve 0.753 end to end accuracy. \looseness=-1
\item To our knowledge, we implement the first attention-based NER for criteria extraction. Our NER detects 10 fine-grained entity classes with precision 0.911 and recall 0.716. \looseness=-1
\item To our knowledge, we implement the first NEL to leverage embedding clustering for criteria grounding. Our NEL achieves an accuracy of 0.485. \looseness=-1
\item We open sourced a portion of the end to end implementation described in this paper, and the largest dataset of clinical trial entities \& attributes which we are aware of.\cref{note1} \looseness=-1
\vspace{-0.2cm}
\end{itemize}
% We open sourced a portion of the IE code and the largest dataset of clinical trial entities and attributes which we are aware of

% The library is not the end to end system described in this paper; rather it is the collection of foundational resources from which our system is derived. 

\section{Dataset Description}

We present a new dataset of 121,221 clinical entities, attributes, and limits taken from 3,314 trials randomly sampled across all disease and treatment areas in ClinicalTrials.gov. Labels were double-annotated by independent layman reviewers with disagreements settled by a senior adjudicator. We define entities as non-parametric patient properties, attributes as numerical/ordinal properties, and limits as constraints on attributes. We believe this dataset represents the largest of its kind; its distribution is shown in Table 1. \looseness=-1

\begin{figure*}
  \centering
  \includegraphics[width=\textwidth]{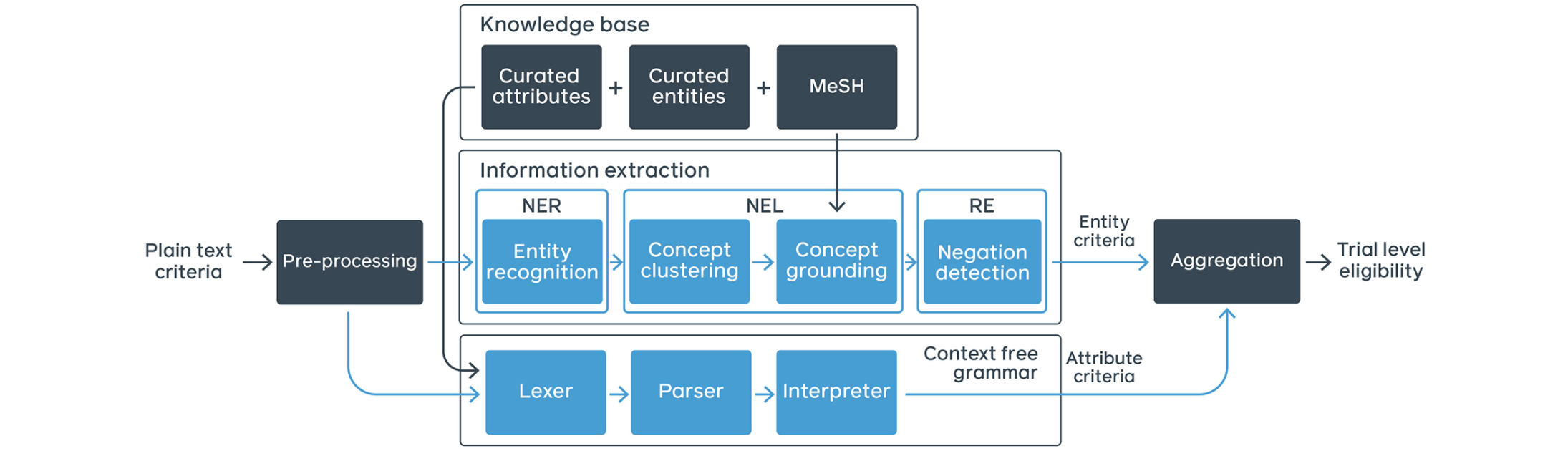}
  \caption{System architecture.}
  \Description{System architecture}
\end{figure*}

\begin{table}
  \centering
  
 \begin{tabular}[t]{ llll } 
  \hline
 & \textbf{Class} & \textbf{Count} & \textbf{Examples} \\
  \hline
Entity & Treatment & 31K & surgery, remdesivir\\
 & Chronic disease & 26K & kidney failure, AD \\
 & Cancer & 9.3K & leukemia \\
 & Gender & 3.7K & ---\\
 & Pregnancy & 2.8K & ---\\
 & Allergy & 1.9K & allergy to aspirin\\
 & Contraception consent & 1.6K & ---\\
 & Language literacy & 482 &  ---\\
 & Technology access & 132 & email, cellphone\\
 & Ethnicity & 82 & ---\\
 \hline
Attribute & Clinical variable & 13K & ECOG, Hgb count\\
& Age & 2.6K & ---\\
& Body mass index & 289 & ---\\
 \hline
Limit & Upper bound & 14K & < 25 $kg/m^{2}$\\
 & Lower bound & 14K & $\geq$ 18 $years$\\
 \hline
\end{tabular}
\Description{Dataset Breakdown}
\caption{Distribution of entity, attribute, and limit classes.}
\vspace{-4mm}
\end{table}

\section{Method}
We divide criteria extraction into two modules: (1) a classic IE pipeline to extract entity criteria and (2) a CFG engine to extract attribute criteria; see Figure 1. 
\looseness=-1

\subsubsection{Criteria Definition}

Given a trial $T$ in ClinicalTrials.org, its criteria text can be split into inclusion criteria blocks ($IC_T$) and exclusion criteria blocks ($EC_T$). $IC_T$ consists of inclusion criteria ($i \in IC_T$) all of which a participant must satisfy in order to be eligible for $T$. $EC_T$ consists of exclusion criteria ($e \in EC_T$) all of which a participant must \textbf{not} satisfy in order to be eligible for $T$. For the purpose of this paper it is assumed all criteria ($c \in IC_T \cup EC_T$) are logically simple. Here a logically simple criterion is defined as a constraint on a single entity or attribute. We formalize the trial-level eligibility for $T$ ($A_T$) as:

\begin{equation}
\label{eq1}
\begin{split}
A_T=IC_T\wedge !EC_T
=\left ( \bigcap^{M}_{x=1} i_x\right) \wedge !\left ( \bigcup^{N}_{x=M+1} e_x\right)
=\bigcap^{N}_{x=1} f(x) \\
f(x) = \left\{\begin{matrix}
i_x, & 1 \leq x \leq M\\ 
!e_x, & M < x \leq N
\end{matrix}\right.
\end{split}
\vspace{-4mm}
\end{equation}

\subsubsection{Task Definition} Extraction of criteria is framed as a classic knowledge base population task. Given unstructured text, the goal is to extract Resource Description Framework (RDF) facts representing individual criterion ($c \in IC_T \cup EC_T$) in the form: 
\begin{equation*}(concept, constraint, trial)\end{equation*}
$concept$ is some entity (e.g. 'leukemia') or attribute (e.g. 'BMI') within the knowledge base. $trial$ is the unique NCT identifier of a trial $T$ in ClinicalTrials.gov (e.g. 'NCT00097734'). $constraint$ is the eligibility requirement $T$ places upon $concept$. Example RDF triplets: ('NCT00097734', 'excludes participants with', 'leukemia'), (`NCT03051984`, 'requires participants have $\leq 38 kg/m^2$', 'BMI').

\subsection{Knowledge Base}
We leverage Medical Subject Headings (MeSH) as our primary knowledge source. MeSH is an NLM controlled vocabulary originally designed as a taxonomy for biomedical research literature. It was chosen for its simplicity and versatility relative to other common knowledge bases such as the International Classification of Diseases (ICD-9/10) and Systematized Nomenclature of Medicine -- Clinical Terms (SNOMED-CT). According to Yao et al.'s analysis, though MeSH (27,000 concepts) is much smaller than ICD-9 (70,000 concepts) and SNOMED-CT (350,000 concepts), it captures the most important disease concepts significantly better relative to both ICD-9 and SNOMED-CT \cite{OntologyComparison}.

We rely on MeSH concepts for treatment, disease, cancer, and allergy entities. We supplement 66 curated concepts for the remaining entity classes (pregnancy, contraception consent, etc), and 71 curated attributes (age, BMI, ECOG, platelet count, etc). Our curated entities and attributes are available on GitHub.\cref{note1} Entities and attributes are organized in a hierarchical structure from general to specific, and seamlessly combine to create one cohesive knowledge base.

\subsection{Pre-Processing}
Trial eligibility is segmented into inclusion and exclusion blocks using rules operating on headings. The text is delexicalized to mask digits and punctuation, normalized for case, and tokenized. Every line of text is then individually considered by the IE and CFG modules.

\begin{table}
\parbox[s]{.49\linewidth}{
\centering
 \begin{tabular}{ ll} 
        \hline
        \multicolumn{2}{c}{Att-BiLSTM-CRF} \\
        \hline
         $\mathbf{Hyper-param.}$ & $\mathbf{Value}$ \\
        batch size & 64 \\
        clipping & $\tau = 1$ \\
        dropout & [0.2, 0.2] \\
        % \hline
        char\_embed dim & 100 \\
        BiLSTM layer & 1 \\
        LSTM dim & 128 \\
        % \hline
        attn dim & 64 \\
        mlp decoder dim & 256\\
          \hline
        \end{tabular}
}
\parbox{.49\linewidth}{
\centering
\begin{tabular}{ ll } 
 \hline
 \multicolumn{2}{c}{Word2vec} \\
 \hline
 $\mathbf{Hyper-param.}$ & $\mathbf{Value}$ \\
model & cbow \\
loss & ns \\
dim & 100 \\
window size & 5 \\
epsilon & $\epsilon = 1.0^{-6}$ \\
learning rate & 0.05 \\
  \hline
\end{tabular}
}
\caption{Hyper-parameter configurations.}
\vspace{-4mm}
\end{table}

\subsection{Entity Criteria Extraction}

\subsubsection{Named Entity Recognition (NER)}
The goal of NER is to extract all entity mentions from unstructured text and categorize mentions by entity class. We trained an attention-based bidirectional Long Short-Term Memory model with a conditional random field layer (Att-BiLSTM-CRF) in PyText \cite{pytorch} for entity recognition of all 10 entity classes. Our model hyper-parameters can be seen in Table 2. Att-BiLSTM-CRF architecture has been shown as state-of-the-art for the task of chemical and disease NER by Luo et al. \cite{Ling18} and Zhai et al. \cite{Zhai18}.

\subsubsection{Named Entity Linking (NEL)} 
The goal of NEL is to link entity mentions with concepts in our knowledge base. We split the task into clustering and grounding. A word2vec model is trained with FastText \cite{mikolov2018advances} on the trial descriptions and eligibility criteria of all 300K+ trials present in ClinicalTrials.gov as of May 2019. Our model hyper-parameters can be seen in Table 2. All extracted entity mentions are projected into the embedding space and clustered with DBSCAN. Clusters are then grounded to entities and their synonyms in the knowledge base according to S{\o}rensen-Dice similarity \cite{Dice}. \looseness=-1

\subsubsection{Relationship Extraction (RE)}
The goal of RE is to identify requirements trials place upon concepts given an unstructured text source. For amenability, we equate the RE task to a simple binary classification of negation detection: Given a trial $T$ and some extracted entity $e$ (e.g. "leukemia"), $T$ must either accept or reject subjects with $e$. This definition of the task does not distinguish temporal requirements: whether $T$ accepts subjects who previously had $e$ or subjects who currently have $e$ is not differentiated.

Medical negation detection is a known problem and solutions usually center around regular expression algorithms \cite{Chapman01}. Our negation detection algorithm is no different. It first searches for specific negation keywords per entity class (e.g. "seronegative" for chronic diseases). Then computes negation according to string distance between keyword and entity mention.

\subsection{Attribute Criteria Extraction}
We rely on a context free grammar (CFG) engine to recognize, ground, and predict criteria for attributes in the knowledge base. A custom lexer divides and categorizes criteria into tokens (attribute, unit, comparison, number, negation, end-of-string, unknown, etc). A modified Cocke-Younger-Kasami (CYK) algorithm builds parse trees from the tokens. The interpreter analyzes the parse trees removing duplicates and sub-trees. The remaining trees are then evaluated into RDF triplets.

\subsection{Aggregation}
Following Equation 1, exclusion criteria are cast to inclusion criteria by negation. To negate RDF facts, the constraint is simply inverted (e.g. "can-have" becomes "cannot-have", ">" becomes "$\leq$"). Inclusion criteria are left unchanged. After casting, all entity and attribute criteria are intersected to calculate trial-level eligibility of $T$ ($A_T$).

Some trials include redundant criteria for clarity; such criteria are de-duplicated during aggregation. Contradicting criteria, which are either the result of an error in our system or a semantic error in the raw criteria text, are also removed. General criteria are dropped in the presence of more specific criteria according to entity and attribute hierarchy; for example ('NCT00594516', 'excludes participants with', 'hepatitis') is dropped because ('NCT00594516', 'excludes participants with', 'hepatitis b') is also extracted. As a final step, constraints that contradict the intent of another constraint or the intent of the trial are removed. For example no trial should require participants to both be pregnant and on birth control.

% We introduce a quality threshold that acts on extracted fact confidence score.

\begin{table*} 
\captionsetup{justification=centering}
  \centering
  
 \begin{tabular}{ llllllll } 
 
 \hline
 & \multicolumn{3}{c}{Our System} &&  \multicolumn{3}{c}{Criteria2Query}\\
 \hline
  
 & \multicolumn{1}{c}{$\mathbf{Precision}$}  & \multicolumn{1}{c}{$\mathbf{Recall}$}  & \multicolumn{1}{c}{$\mathbf{F1}$} &      & \multicolumn{1}{c}{$\mathbf{Precision}$}  & \multicolumn{1}{c}{$\mathbf{Recall}$}  & \multicolumn{1}{c}{$\mathbf{F1}$}  \\
  Entity Recognition & 
  \multicolumn{1}{c}{\begin{tabular}{@{}c@{}}0.911 (154/169)\\{[0.864-0.953]}\end{tabular}} & 
  \multicolumn{1}{c}{\begin{tabular}{@{}c@{}}0.716 (154/215)\\{[0.656-0.772]}\end{tabular}} & 
  \multicolumn{1}{c}{\begin{tabular}{@{}c@{}}0.802\\{[0.754-0.837]}\end{tabular}} && \multicolumn{1}{c}{\begin{tabular}{@{}c@{}}0.902 (156/173)\\{[0.844-0.936]}\end{tabular}} & \multicolumn{1}{c}{\begin{tabular}{@{}c@{}}0.726 (156/215)\\{[0.661-0.777]}\end{tabular}} & \multicolumn{1}{c}{\begin{tabular}{@{}c@{}}0.804\\{[0.760-0.841]}\end{tabular}} \\
  \hline
  &\multicolumn{3}{c}{$\mathbf{Accuracy}$}&&\multicolumn{3}{c}{$\mathbf{Accuracy}$} \\
  Entity Linking & \multicolumn{3}{c}{0.485 (82/169) {[0.408-0.556]}} && \multicolumn{3}{c}{0.447 (51/114) {[0.351-0.535]}}\\
  Attribute Linking & \multicolumn{3}{c}{0.750 (15/20) {[0.450-0.850]$^\star$}}&&\multicolumn{3}{c}{0.800 (16/20) {[0.500-0.900]}}\\
  Relationship Extraction  & \multicolumn{3}{c}{0.838 (57/68) {[0.750-0.926]}}&&\multicolumn{3}{c}{-}\\
  End to End Performance  & \multicolumn{3}{c}{0.753 (64/85) {[0.661-0.844]}}&&\multicolumn{3}{c}{-}\\
  \hline
\end{tabular}
\Description{Results on golden set.}
\caption{Evaluation on Criteria2Query's golden set with 95\% confidence interval.\\ $^{\star}$Evaluation of this metric is inferred; refer to section 4.0.3 for more detail.}
\vspace{-1mm}
\end{table*}

\section{Results \& Discussion}
Table 3 shows the performance of our system as evaluated on the 10 trial golden set created by Yuan et al. for Criteria2Query \cite{Criteria2Query}. We release the evaluations from our internal build as ancillary files.\footnote{\href{https://arxiv.org/abs/2006.07296}{https://arxiv.org/abs/2006.07296}} \looseness=-1

\subsubsection{Entity Recognition} Our NER model employs an Att-BiLSTM-CRF architecture trained with data from 3,314 randomly sampled trials. It predicts 10 fine-grained entity classes shown in Table 1 at 0.802 F1 score. Criteria2Query's NER model employs a classic CRF architecture trained with data from 230 Alzheimer’s disease trials. It predicts 5 general entity classes (Condition, Drug, Measurement, Procedure, Observation) at 0.804 F1 score \cite{Criteria2Query}. Our NER model extracts more fine-grained entities while maintaining a competitive F1 score. The Att-BiLSTM-CRF architecture has been successfully applied to many medical applications \cite{Ling18}\cite{Zhai18}. To our knowledge, this work is the first to apply Att-BiLSTM-CRF, or any other attention-based architecture, to clinical trial criteria extraction. Investigating other promising architectures such as BioBERT \cite{BioBERT} for clinical trial NER is a direction for future work.

\subsubsection{Entity Linking} Our NEL accuracy of 0.485 (82/169) slightly outperforms the 0.447 (51/114) accuracy of Criteria2Query's NEL module. Criteria2Query was intended as a companion tool for trial investigators. For NEL, it relies on predefined concept sets augmented by a sophisticated interface for creating new custom concept sets \cite{Criteria2Query}. Our NEL took a fundamentally different approach. We trained a word2vec model from trial descriptions which we then used to project, cluster, and ground entity embeddings to our knowledge base. In this way our NEL construction was self-supervised, and can be easily configured to ground any knowledge base. \looseness=-1

From an absolute perspective, NEL is the bottleneck of our pipeline. It fails to ground 0.416 (64/154) of valid extractions. Of incorrect groundings, 0.696 (16/23) are from over generalization (e.g. "left main stem stenosis" is incorrectly grounded to "stenosis"). We believe these limitations are the consequence of our knowledge base, composed primarily of MeSH concepts. MeSH is comprised of orders of magnitude fewer and more general concepts relative to other standards such as SNOMED-CT and ICD-9/10 \cite{OntologyComparison}. In the task of trial criteria extraction, requirements can be exceedingly specific \cite{Ross10}. To accurately link entities requires a comprehensive knowledge base of equivalently specific concepts. We leave experimentation with specialized and expanded knowledge bases (e.g. curated breast cancer trial requirements, SNOMED-CT) for future work. \looseness=-1

\subsubsection{Attribute Linking} Attribute recognition, linking and constraint extraction are performed in tandem by our CFG engine. Evaluated on the golden set, the end to end precision of our attribute criteria extraction is 0.938 (15/16). Criteria2Query similarly relies upon rules for attribute extraction, but does not report its end to end accuracy, only its linking accuracy of 0.800 (16/20) \cite{Criteria2Query}. For the purpose of comparison, we infer the attribute linking accuracy of our CFG engine to be 0.750 (15/20). \looseness=-1

\subsubsection{Relationship Extraction} We achieve an accuracy of 0.838 (57/68) for our rule-based negation detection module. There is no counterpart to our RE module in Criteria2Query. Criteria2Query frames its RE task to infer relations between entities and attributes, rather than concepts and trials. 

The expressiveness of our framework is constrained by the simplification of RE to just binary negation detection, and our assumption that all criteria are logically simple. Collectively, 0.636 (7/11) of RE errors could be avoided by expanding the relation ontology to capture a wider array of requirements (e.g. history of disease), and handling logically complex criteria (e.g. conditional and compound constraints). \looseness=-1

\subsubsection{End to End Performance} Our end to end accuracy on Criteria2Query's golden set is 0.753 (64/85) as evaluated by a medical professional. Criteria2Query does not report its end to end accuracy for comparison. However, EliIE, the precursor to Criteria2Query, does report its end to end accuracy of 0.71 \cite{EliIE}.

\vspace{3mm}

\section{Conclusion}

We present a novel formulation of clinical trial eligibility criteria extraction as a knowledge base population task. As far as we are aware, this work is the first to apply attention-based architecture to clinical trial entity extraction, and word2vec embedding clustering to clinical trial entity linking. We open source a library containing our training data, CFG, embeddings, and NER model binary. We evaluate our system against Yuan et al.'s Criteria2Query pipeline \cite{Criteria2Query}, which we consider the state-of-the-art, to demonstrate the competitiveness of our system.

\small
\bibliographystyle{abbrv}
\bibliography{references}

\begin{thebibliography}{10}

\bibitem{Alves2019InformationEA}
S.~G. Alves, J.~S. Costa, and J.~Bernardino.
\newblock Information extraction applications for clinical trials: A survey.
\newblock {\em 2019 14th Iberian Conference on Information Systems and
  Technologies (CISTI)}, pages 1--6, 2019.

\bibitem{EliXR-TIME}
M.~R. Boland, S.~W. Tu, S.~Carini, I.~Sim, and C.~Weng.
\newblock {{E}li{X}{R}-{T}{I}{M}{E}: {A} {T}emporal {K}nowledge
  {R}epresentation for {C}linical {R}esearch {E}ligibility {C}riteria}.
\newblock {\em AMIA Jt Summits Transl Sci Proc}, 2012:71--80, 2012.

\bibitem{Butler18}
A.~Butler, W.~Wei, C.~Yuan, T.~Kang, Y.~Si, and C.~Weng.
\newblock {{T}he data gap in {E}{H}{R} for clinical research eligibility
  screening}.
\newblock {\em AMIA Jt Summits Transl Sci Proc}, 2018.

\bibitem{Carlisle05}
B.~Carlisle, J.~Kimmelman, T.~Ramsay, and N.~MacKinnon.
\newblock {{U}nsuccessful trial accrual and human subjects protections: an
  empirical analysis of recently closed trials}.
\newblock {\em Clin Trials}, 12(1):77--83, Feb 2015.

\bibitem{Chapman01}
W.~Chapman, W.~Bridewell, P.~Hanbury, G.~F. Cooper, and B.~G. Buchanan.
\newblock {A Simple Algorithm for Identifying Negated Findings and Diseases in
  Discharge Summaries}.
\newblock {\em Journal of Biomedical Informatics}, 34(5):301--310, Oct 2001.

\bibitem{Bruijn08}
B.~de~Bruijn, S.~Carini, S.~Kiritchenko, J.~Martin, and I.~Sim.
\newblock {Automated information extraction of key trial design elements from
  clinical trial publications}.
\newblock {\em {AMIA Annu Symp Proc}}, pages 141--145, 2008.

\bibitem{Dice}
L.~R. Dice.
\newblock Measures of the amount of ecologic association between species.
\newblock {\em Ecology}, 26(3):297--302, 1945.

\bibitem{Gubar19}
S.~Gubar.
\newblock {The Need for Clinical Trial Navigators}.
\newblock {\em The New York Times}, Jun 2019.

\bibitem{EliIE}
T.~Kang, S.~Zhang, Y.~Tang, G.~Hruby, A.~Rusanov, N.~Elhadad, and C.~Weng.
\newblock {E}li{IE}: An open-source information extraction system for clinical
  trial eligibility criteria.
\newblock {\em JAMIA}, 24(6):1062--1071, Nov 2017.

\bibitem{Knepper18}
T.~Knepper and H.~McLeod.
\newblock {When will clinical trials finally reflect diversity?}
\newblock {\em Scientific American}, (557):157--159, May 2018.

\bibitem{BioBERT}
J.~Lee, W.~Yoon, S.~Kim, D.~Kim, S.~Kim, C.~H. So, and J.~Kang.
\newblock Biobert: a pre-trained biomedical language representation model for
  biomedical text mining.
\newblock 2019.

\bibitem{Ling18}
L.~Luo, Z.~Yang, P.~Yang, Y.~Zhang, L.~Wang, H.~Lin, and J.~Wang.
\newblock {An attention-based BiLSTM-CRF approach to document-level chemical
  named entity recognition }.
\newblock {\em Bioinformatics}, 32(8):1381--1388, Apr 2018.

\bibitem{Luo13}
Z.~Luo, R.~Miotto, and C.~Weng.
\newblock {{A} human-computer collaborative approach to identifying common data
  elements in clinical trial eligibility criteria}.
\newblock {\em J Biomed Inform}, 46(1):33--39, Feb 2013.

\bibitem{mikolov2018advances}
T.~Mikolov, E.~Grave, P.~Bojanowski, C.~Puhrsch, and A.~Joulin.
\newblock Advances in pre-training distributed word representations.
\newblock In {\em Proceedings of the International Conference on Language
  Resources and Evaluation (LREC 2018)}, 2018.

\bibitem{pytorch}
A.~Paszke, S.~Gross, S.~Chintala, G.~Chanan, E.~Yang, Z.~DeVito, Z.~Lin,
  A.~Desmaison, L.~Antiga, and A.~Lerer.
\newblock Automatic differentiation in pytorch.
\newblock 2017.

\bibitem{Ross10}
J.~Ross, S.~Tu, S.~Carini, and I.~Sim.
\newblock {{A}nalysis of eligibility criteria complexity in clinical trials}.
\newblock {\em Summit Transl Bioinform}, 2010:46--50, Mar 2010.

\bibitem{ERGO}
S.~W. Tu, M.~Peleg, S.~Carini, M.~Bobak, J.~Ross, D.~Rubin, and I.~Sim.
\newblock {{A} practical method for transforming free-text eligibility criteria
  into computable criteria}.
\newblock {\em J Biomed Inform}, 44(2):239--250, Apr 2011.

\bibitem{Weng10}
C.~Weng, S.~W. Tu, I.~Sim, and R.~Richesson.
\newblock {{F}ormal representation of eligibility criteria: a literature
  review}.
\newblock {\em J Biomed Inform}, 43(3):451--467, Jun 2010.

\bibitem{EliXR}
C.~Weng, X.~Wu, Z.~Luo, M.~R. Boland, D.~Theodoratos, and S.~B. Johnson.
\newblock {{E}li{X}{R}: an approach to eligibility criteria extraction and
  representation}.
\newblock {\em J Am Med Inform Assoc}, 18 Suppl 1:i116--124, Dec 2011.

\bibitem{OntologyComparison}
L.~Yao, A.~Divoli, I.~Mayzus, J.~A. Evans, and A.~Rzhetsky.
\newblock {Benchmarking ontologies: bigger or better?}
\newblock {\em {PLoS Comput Biol}}, 7(1), 2011.

\bibitem{Criteria2Query}
C.~Yuan, P.~B. Ryan, C.~Ta, Y.~Guo, Z.~Li, J.~Hardin, R.~Makadia, P.~Jin,
  N.~Shang, T.~Kang, and C.~Weng.
\newblock {C}riteria2{Q}uery: a natural language interface to clinical
  databases for cohort definition.
\newblock {\em JAMIA}, 26(4):294--305, Apr 2019.

\bibitem{ClinicalTrialsdotgov}
D.~A. Zarin, T.~Tse, R.~J. Williams, R.~M. Califf, and N.~C. Ide.
\newblock {The ClinicalTrials.gov results database--update and key issues}.
\newblock {\em {N Engl J Med}}, 364(9):852--860, 2011.

\bibitem{Zhai18}
Z.~Zhai, D.~Q. Nguyen, and K.~Verspoor.
\newblock {Comparing CNN and LSTM character-level embeddings in BiLSTM-CRF
  models for chemical and disease named entity recognition}.
\newblock {\em arxiv}, Aug 2018.

\end{thebibliography}

\end{document}